\documentclass[letterpaper, 10 pt, journal, twoside]{IEEEtran}
\usepackage{cite}

\ifCLASSINFOpdf
  \usepackage[pdftex]{graphicx}
\else
\fi
\usepackage{amsmath}

\usepackage{multirow}

\begin{document}
%
\title{Time-Relative RTK-GNSS:\\GNSS Loop Closure in Pose Graph Optimization}
%
%
%

\author{Taro Suzuki%
\thanks{Manuscript received: February 24, 2020; Revised May 21, 2020; Accepted June 15, 2020. This letterwas recommended for publication by Associate Editor S. Behnke and N. Atanasov upon evaluation of the reviewers’ comments.}
\thanks{The author is with the Future Robotics Technology Center, Chiba Institute of Technology, Narashino-shi 2750016, Japan (e-mail: taro@furo.org).}%
\thanks{Digital Object Identifier 10.1109/LRA.2020.3003861}
}
%
%

\markboth{IEEE Robotics and Automation Letters. Preprint Version. Accepted June, 2020}
{Taro Suzuki: Time-Relative RTK-GNSS: GNSS Loop Closure in Pose Graph Optimization} 

%



\maketitle

\begin{abstract}
  A pose-graph-based optimization technique is widely used to estimate robot poses using various sensor measurements from devices such as laser scanners and cameras. The global navigation satellite system (GNSS) has recently been used to estimate the absolute 3D position of outdoor mobile robots. However, since the accuracy of GNSS single-point positioning is only a few meters, the GNSS is not used for the loop closure of a pose graph. The main purpose of this study is to generate a loop closure of a pose graph using a time-relative real-time kinematic GNSS (TR-RTK-GNSS) technique. The proposed TR-RTK-GNSS technique uses time–differential carrier phase positioning, which is based on carrier-phase-based differential GNSS with a single GNSS receiver. Unlike a conventional RTK-GNSS, we can directly compute the robot’s relative position using only a stand-alone GNSS receiver. The initial pose graph is generated from the accumulated velocity computed from GNSS Doppler measurements. To reduce the accumulated error of velocity, we use the TR-RTK-GNSS technique for the loop closure in the graph-based optimization framework. The kinematic positioning tests were performed using an unmanned aerial vehicle to confirm the effectiveness of the proposed technique. From the tests, we can estimate the vehicle's trajectory with approximately 3 cm accuracy using only a stand-alone GNSS receiver.
\end{abstract}


%
\IEEEpeerreviewmaketitle

\section{Introduction}
%
%
%
%
\IEEEPARstart{G}{lobal} navigation satellite system (GNSS) has become an important component in self-driving cars and outdoor mobile robots to estimate the global location. However, GNSS data are not utilized to their full potential in autonomously navigating vehicles. A 3D absolute position in geospatial coordinates can be estimated using the GNSS. However, the accuracy of GNSS single-point positioning is only a few meters, which is insufficient for autonomous navigation. Real-time kinematic (RTK)GNSS positioning technology can be used to estimate the position with centimeter-level accuracy. However, one disadvantage of RTK-GNSS is that, it requires an additional GNSS base station to estimate the precise position. In addition, the RTK-GNSS requires a communication link to obtain the correction data from a GNSS base station. 

To overcome these issues, we leverage the advances made within the robotics community related to pose-graph-based optimization and simultaneous localization and mapping (SLAM) techniques \cite{slam1,slam2,slam3}. One intuitive method of formulating SLAM involves using a graph with nodes corresponding to the various poses of the robot at different points in time and edges representing the constraints between the poses. After the graph is constructed, it is optimized and the poses of the robot can be estimated.

\begin{figure}[t]
   \centering
   \includegraphics[width=85mm]{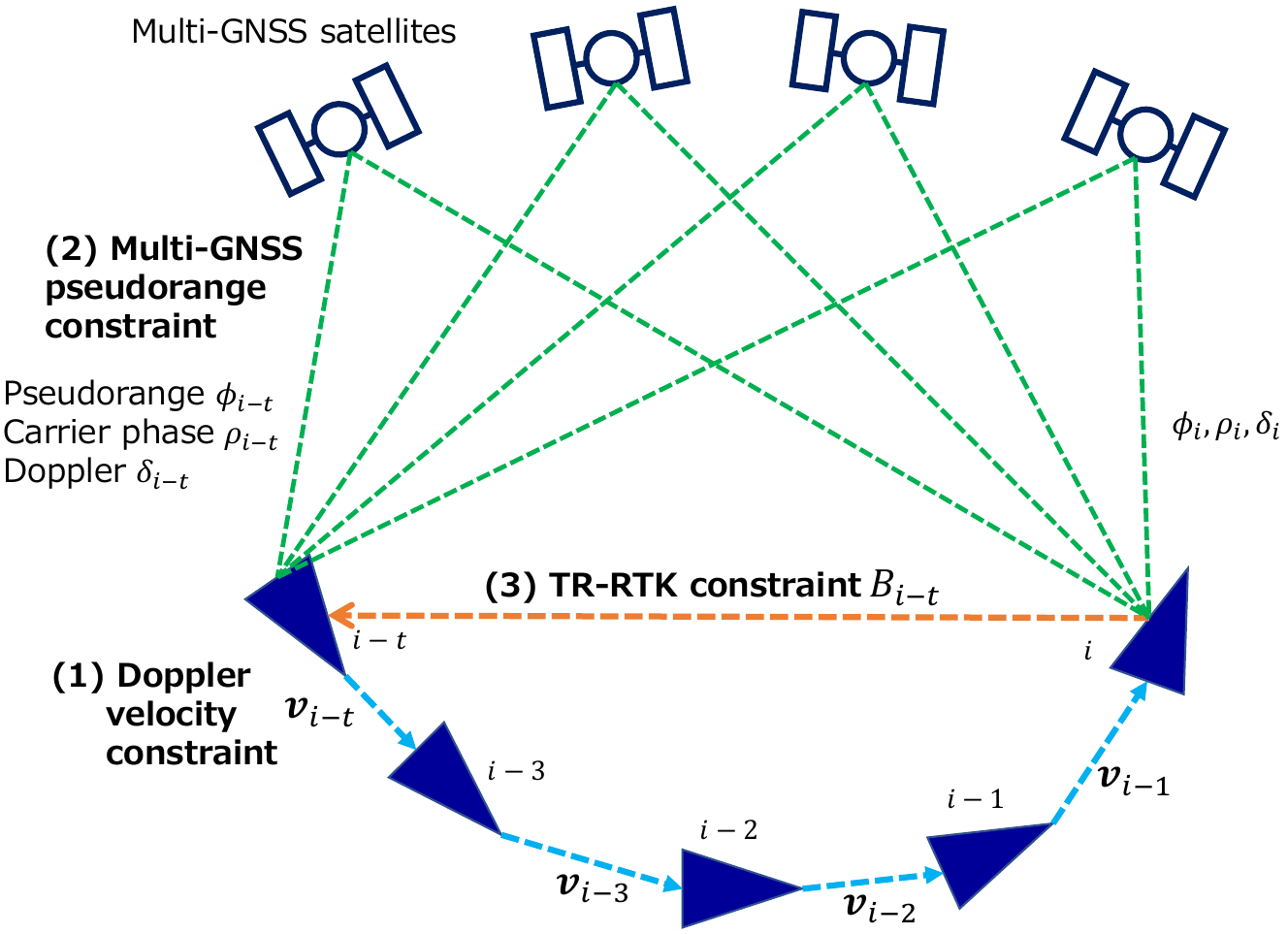} 
   \caption{Concept of the proposed graph optimization technique. We use three constraints: (1) Doppler velocity constraints between successive nodes, (2) multi-GNSS pseudorange constraints to estimate global position, and (3) novel TR-RTK-GNSS constraints to reduce the accumulated error of the Doppler velocity.}
   \label{fig1}
\end{figure}

The main purpose of this study is to generate a loop closure of a pose graph from a time-relative (TR) RTK-GNSS technique. In conventional methods, GNSS pseudorange measurements are used to estimate the absolute position in a pose graph. In contrast, we use the edge between the distant nodes computed from TR-RTK-GNSS to estimate the absolute position in centimeter-level accuracy. Fig. 1 shows the loop closing of a pose graph to reduce the accumulated error. The loop-closing of the pose graph is critical to reduce the accumulated error based on a relative constraint such as velocity. The proposed TR-RTK-GNSS technique uses time–differential carrier phase positioning, which is based on carrier-phase-based differential GNSS with a single GNSS receiver. In the TR-RTK-GNSS technique, previous GNSS pseudorange and carrier phase measurements are stored and used to construct double-difference measurements between the past and current measurements. In contrast to a conventional RTK-GNSS technique that uses an additional base station, in the TR-RTK-GNSS technique, a stand-alone GNSS receiver is sufficient to directly compute the vehicle's relative position in centimeter-level accuracy. Another characteristic of the proposed method is that neither inertial measurement units (IMUs) nor odometry is used to construct the relative constraints; instead, a GNSS Doppler velocity is used. Thus, the proposed method does not use any additional sensors. Further, pseudorange measurements of a multi-GNSS constellation can be used by simultaneously estimating the GNSS inter systems clock biases. The initial pose graph is generated from the accumulated velocity computed from GNSS Doppler measurements. To correct the accumulated error of velocity, we can use the TR-RTK-GNSS for the loop closure in the graph-based optimization framework. The proposed method uses only a stand-alone GNSS receiver to construct the graph without employing any other sensors.

\subsection{Related Works}

Precise velocity measurement is required to construct the relative constraint in graph-based optimization. An IMU, wheel odometry, camera, or laser scanner are typically used to estimate the relative constraint \cite{slam1, slam3}. In this study, we use GNSS Doppler measurement. Doppler-based velocity determination is a widely used technique and usually has a cm/s accuracy, whereas the velocity obtained by differencing consecutive positions obtained from single-point positioning has an accuracy of one order of magnitude greater. The precise Doppler-based velocity obtained is used to estimate vehicle trajectories in many studies \cite{vel2,vel3}. However, in these studies the focus is on estimating the precise velocity and no optimization techniques are used to estimate the vehicle trajectory.
The TR-RTK-GNSS is a novel approach that can be used for loop closures in graph optimizations. In the previous studies, GNSS was never used as the loop closure in pose-graph optimizations. Many techniques have been proposed to eliminate the accumulated error by using a camera, laser scanner, or other sensors \cite{slam1,slam2,slam3}. However, no methods exist for estimating the loop-closing edges using GNSS. The time-differenced carrier phase technique was proposed to estimate the accurate velocity and delta position \cite{td1,td2,td3}. However, the velocity and delta position were directly computed from the GNSS carrier phase and they were not used in optimization frameworks.
Many studies have focused on applying factor graphs as a tool for least-squares optimization in SLAM problems. A broad field of classic sensor fusion problems that were previously addressed using filter approaches have benefitted from formulating them as graph optimization problems. Other works related to this study include \cite{gognss0,gognss1,gognss2,gognss3,gognss4},and these literatures propose robust graph optimization for GPS-based applications. They use pseudorange constraints for graph-optimization and do not consider loop-closing edges. In addition, these works only focus on GPS pseudorange; multi-GNSS constellations such as GLONASS, Galileo, and BeiDOU are not considered.

\subsection{Contributions}
The contributions of this study are as follows. 
\begin{itemize}
\item To the best of our knowledge, this paper is the first paper to use GNSS as a loop closure in pose-graph optimization. 
\item We propose a novel graph-based localization method using TR-RTK-GNSS for a loop closure, and the proposed method can be used to estimate vehicle trajectory with centimeter accuracy and without using a GNSS base station or other sensors. 
\item We also propose error functions for graph-based optimization using a multi-GNSS constellation. 
\end{itemize}

\section{Factor Graph}

\begin{figure*}[htpb]
   \centering
   \includegraphics[width=165mm]{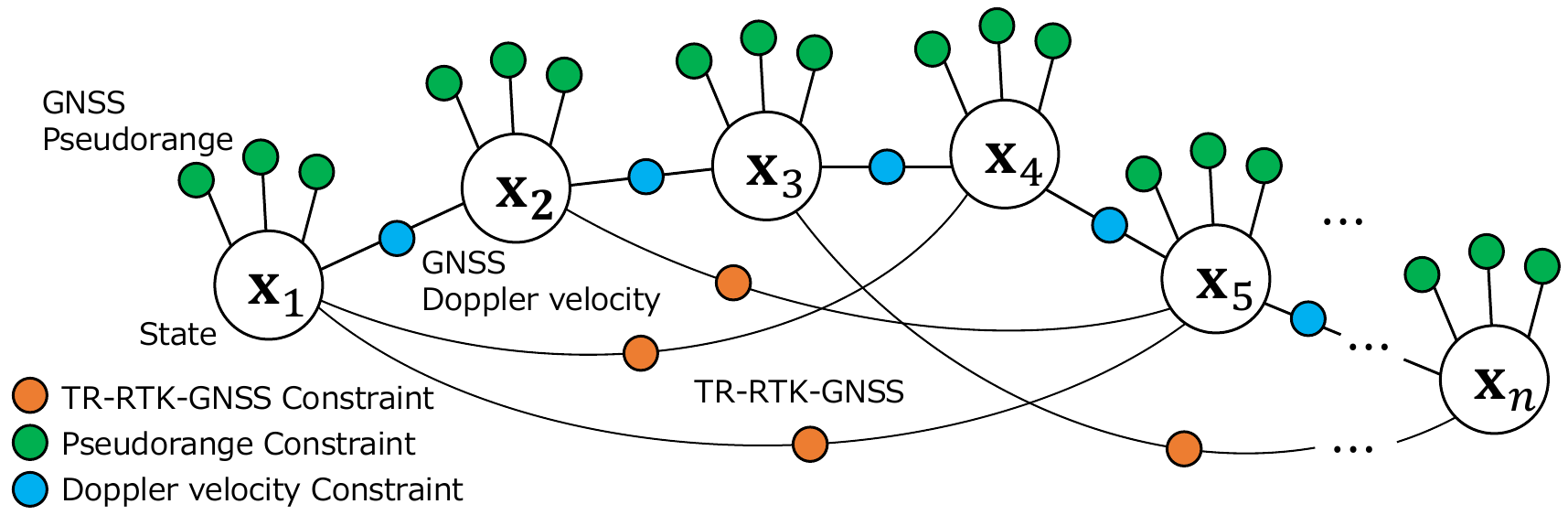} 
   \caption{Graph structure of the proposed method. The node $X$ consists of the 3D position and each of the GNSS clock bias. The loop-closing edge is computed from the TR-RTK-GNSS technique.}
   \label{fig2}
\end{figure*}

The use of a factor graph was proposed in \cite{go1} to model factorizations. Formally, a factor graph is a bipartite undirected graph that includes two types of nodes, namely variable nodes and factor nodes, and edges. A variable node presents the state variable ${X}$ at different moments. The edge connects the factor node and variable node and encodes the error function ${e}(\cdot)$, in which each edge corresponds to a single observation ${Z}$. The error function ${e}(\cdot)$ represents the probabilistic constraints applied to a state at a specified time step. The aim is to find the optimal state $\widehat{X}$ that minimizes the sum of all error functions. The optimization factor graph can be represented as: 

\begin{equation}
\widehat{{X}}=\underset{{X}}{\operatorname{argmin}} \sum_{i}\left\|{e}\left({X}_{i}, {Z}_{i}\right)\right\|_{\Omega_{i}}^{2}
\end{equation}

where $\Omega_{i}$ represents the information matrix of the constraint and determines the weight of error function over the entire optimization problem. Estimation of the optimal state can be calculated by using a traditional non-linear least-squares formulation that minimizes the error over the graph. This is a least-squares optimization problem because it seeks the minimum over a sum of squared terms. These types of problems can be solved using a variety of methods such as the Levenberg-Marquardt, Gauss-Newton, or Powell's Dog-Leg \cite{tuto}. In these approaches, the problem is iteratively solved by repeatedly linearizing it and updating the current estimate of $\widehat{{X}}$ until convergence is reached.

\section{Proposed Method}

\subsection{Overview}

In this study, we estimate the vehicle position using the factor graph. GNSS pseudorange, carrier phase, and Doppler measurements are used as the constraints of the factor graph. The state ${X}$ estimated in epoch $i$ is the 3D position relative to the start position in earth center earth fixed (ECEF) coordinates and multi-GNSS clock biases. The state ${X}$ can be expressed as:

\begin{equation}
{X}_{i}=\left[\begin{array}{lllllll}
{d_{x, i}} & {d_{y, i}} & {d_{z, i}} & {d_{\mathrm{GPS}, i}} & {d_{\mathrm{GLO}, i}} & {d_{\mathrm{GAL}, i}} & {d_{\mathrm{BDS}, i}}
\end{array}\right]^{T}
\end{equation}

Fig. 2 illustrates the basic concept of graph-based optimization using the TR-RTK-GNSS technique. Three main factors exist: relative state, multi-GNSS pseudorange, and TR-RTK-GNSS. $n$ is the total number of estimated states. First, we estimate the velocity of the vehicle using GNSS Doppler measurements. The graph nodes are then generated by the accumulated velocities. The blue circle in Fig. 2 represents the relative state factors computed based on the Doppler velocity. The orange circles that constitute loop-closing edges represent the TR-RTK-GNSS factors. The green circle represents the multi-GNSS pseudorange factor that functions as the global constraint to estimate positions in geospatial coordinates. The accumulated error of the Doppler velocity between states is reduced by the TR-RTK-GNSS constraints to optimize this entire graph using the non-linear least-squares technique. Finally, an accurate vehicle trajectory can be estimated using only a stand-alone GNSS receiver and without other sensors. 

In the next section, the details of the relative state constraints computed based on the Doppler velocity, the TR-RTK-GNSS constraints, and multi-GNSS pseudorange constraints are described.

\subsection{Doppler Velocity Constraint}

The GNSS Doppler frequency can be used to derive an accurate velocity without the need of a base station. This velocity is tolerant to the strong multipath condition \cite{vel1} as compared to the GNSS pseudorange-based position solution. Thus, the Doppler velocity can be used to estimate vehicle trajectories \cite{vel2, vel3}. However, positioning errors are accumulated over time as a result of the error in the estimated Doppler velocity, and therefore, the velocity estimated from a Doppler cannot be used alone. In some studies, an inertial navigation system was used \cite{vel2}, or the RTK-GNSS was combined with the Doppler velocity \cite{vel3} to estimate the accurate vehicle position. However, issues such as system complexity and cost negatively impact their use.

To estimate the vehicle trajectory, first, the velocity information is used to estimate the relative position. The velocity is used as the constraint between the time sequential states in the graph. The stand-alone GNSS receiver can be used to estimate the velocity by using Doppler measurements relative to the user-satellite motion. Doppler-based velocity is the most widely used technique and usually has a cm/s accuracy, whereas the velocity obtained by differencing consecutive positions from single-point positioning has an accuracy of one order of magnitude greater. The absolute 3D velocity is estimated based on the Doppler measurements using the conventional least squares technique. The velocity estimated from the Doppler measurement is denoted as ${V}_{i}=\left[\begin{array}{lll}{v_{x, i}} & {v_{y, i}} & {v_{z, i}}\end{array}\right]^{T}$. The error function of the relative state constraint is:

\begin{equation}
\boldsymbol{e}_{\mathrm{vel}, i}=\left[\begin{array}{l}{d_{x, i+1}-d_{x, i}} \\ {d_{y, i+1}-d_{y, i}} \\ {d_{z, i+1}-d_{z, i}}\end{array}\right]-V_{i} \cdot \Delta t
\end{equation}

where $\Delta t$ is the time step of the GNSS observations. The minimized error is calculated as:

\begin{equation}
\left\|e_{\mathrm{vel}, i}\right\|_{\Omega_{\mathrm{vel}, i}}=e_{\mathrm{vel}, i}\hspace{1pt} \Omega_{\mathrm{vel},i} \hspace{1pt} e_{\mathrm{vel}, i}
\end{equation}

where $\Omega_{\mathrm{vel}}$ represents the information matrix, which is the inverse of the covariance matrix; it indicates the uncertainty of velocity. It should be noted that, the velocity estimated from the Doppler measurement is also affected by the multipath. However, this velocity is quite tolerant to the strong multipath condition as compared to the GNSS pseudorange-based position solution. 

\subsection{TR-RTK-GNSS Constraint}

This study attempts to construct loop-closure edges using TR-RTK-GNSS technique to reduce the accumulated trajectory error by relative state constraints. Previous GNSS pseudorange and carrier phase measurements are stored and used to construct double-difference measurements between the past and current measurements. The subsequent trajectory is determined by the time-differential processing of RTK-GNSS. With respect to other terms, the base vectors pointing from the past epoch to the current position are determined exactly. We use these base vectors as the loop-closure edges in the graph. The GNSS carrier phase measurement of satellite $k$ in epoch $i$ is denoted as $\Phi_{i}^{k}$. $\Phi_{i}^{k}$ is given by:

\begin{equation}
\Phi_{i}^{k}=r_{i}^{k}+\lambda N_{i}^{k}+c\left(\delta t_{i}-\delta T_{i}^{k}\right)-I_{i}^{k}+T_{i}^{k}+\epsilon_{i}^{k}
\end{equation}

where $r_{i}^{k}$ is the range from the receiver to the satellite $k$, $\lambda$ is the signal wavelength, $N_{i}^{k}$ is the integer ambiguity, $c$ is the speed of light, $\delta t_{i}$ and $\delta T_{i}^{k}$ are the receiver and satellite clock biases, respectively, $I_{i}^{k}$and $T_{i}^{k}$ are the ionosphere and troposphere errors, and $\epsilon_{i}^{k}$ is the carrier phase measurement noise. In general, when two receivers are used, mutual error terms exist in the measurements from the same satellite. The ionosphere, troposphere, and satellite clock bias errors may be eliminated by differencing the base and rover receiver measurements. In contrast, we consider the time-differential carrier phase using a stand-alone GNSS receiver. The time-differential carrier phase between epoch $i$ and past epoch $i-t$ is given as:

\begin{equation}
\begin{aligned} \Delta \Phi_{i, i-t}^{k} &=\Phi_{i}^{k}-\Phi_{i-t}^{k} \\ &\approx \Delta r_{i, i-t}^{k}+\lambda \Delta N_{i, i-t}^{k}+c\left(\Delta \delta t_{i, i-t}\right)+\Delta \epsilon_{i, i-t}^{k} \end{aligned}
\end{equation}

where $t$ is the time from the current epoch. In the time-differential process, the ionosphere and troposphere delays are not rapidly changed and therefore, can be canceled. Regarding the satellite clock bias, it is changed more rapidly than the ionosphere and troposphere delays. However, it is practically negligible if the duration of the time difference $t$ is less than approximately 1 or 2 min. In Equation (6), we eliminated the satellite clock bias $\delta T_{i}^{k}$ under the assumption that the duration of time difference $t$ is not large. The GNSS carrier phase measurements are biased by an unknown integer ambiguity. This ambiguity will be a time-invariant constant, provided that there is a continuous phase lock to the respective GNSS satellite in the receiver phase lock loop. However, the GNSS carrier phase measurements are sensitive to signal shadowing. In urban environments, tall buildings and overpasses are likely to cause complete signal obstruction. In these situations, cycle slip of the GNSS carrier phase occurs, and the integer ambiguity will not be a time-invariant constant. For this reason, we do not eliminate integer ambiguity in the time difference computation in Equation (6).

The measurement differencing technique can again be applied to eliminate the receiver clock bias error $\Delta \delta t_{i, i-t}$ by differencing the measurements between the two satellites $k$ and $l$. This is called as double differencing (DD) and is given as:

\begin{equation}
\begin{aligned} \nabla \Delta \Phi_{i, i-t}^{k, l} &=\Delta \Phi_{i, i-t}^{k}-\Delta \Phi_{i, i-t}^{l} \\ &=\nabla \Delta r_{i, i-t}^{k, l}+\lambda \nabla \Delta N_{i, i-t}^{k, l}+\nabla \Delta \epsilon_{i, i-t}^{k, l} \end{aligned}
\end{equation}

The DD measurements are the sum of the DD ranges and DD integer ambiguities with additive noise. In the TR-RTK-GNSS, the baseline vector, which is a relative 3D vector from the past position to the current position, is denoted as ${B}_{i-t}=\left[\begin{array}{lll}{b_{x, i-t}} & {b_{y, i-t}} & {b_{z, i-t}}\end{array}\right]^{T}$. The relationship between the DD range $\nabla \Delta r_{i, i-t}^{k, l}$ and the baseline vector ${B}_{i-t}$ is represented as:

\begin{equation}
   \nabla \Delta r_{i, i-t}^{k, l}=-\left( L_{i}^{k}-L_{i}^{l} \right)^T{B}_{i-t}
\end{equation}
   
where $L_{i}^{k}$ and $L_{i}^{l}$ are the normalized line-of-sight vectors pointing from the position of the current epoch to the satellites $k$ and $l$. To estimate the baseline vector ${B}_{i-t}$, the DD integer ambiguities $\nabla \Delta N_{i, i-t}^{k, l}$ must be estimated simultaneously. The form of the observed equation of TR-RTK-GNSS shown in Equation (7) is similar to that of normal RTK-GNSS. Therefore, the baseline vectors and ambiguity are estimated using the ambiguity estimation method used in RTK-GNSS. In general, the ambiguities are first fixed to float numbers using a Kalman filter or a least-squares method from the DD carrier phase measurements in single or multiple epoch. Then float ambiguities are corrected with the integer ambiguities, and the accurate baseline vector can be computed using the estimated integer ambiguities. We use the LAMBDA method \cite{lambda} to estimate the integer ambiguities, and the precise relative position between the past and current epochs can be estimated. The reference \cite{rtklib} provides a more specific procedure for ambiguity estimation.

The error function of the TR-RTK-GNSS constraint is:

\begin{equation}
\boldsymbol{e}_{\mathrm{tr}}=\left[\begin{array}{l}{d_{x, i}-d_{x, i-t}} \\ {d_{y, i}-d_{y, i-t}} \\ {d_{z, i}-d_{z, i-t}}\end{array}\right]-{B}_{i-t}
\end{equation}

The minimized error is calculated as:

\begin{equation}
\left\|e_{\mathrm{tr}, j}\right\|=e_{\mathrm{tr}, j} \hspace{1pt} \Omega_{\mathrm{tr}, j} \hspace{1pt} e_{\mathrm{tr}, j}
\end{equation}

If TR-RTK-GNSS can correctly estimate the carrier phase ambiguity as an integer, we add all the TR-RTK-GNSS constraints to the graph. We do not add the constraint if the integer carrier phase ambiguity cannot be solved by the LAMBDA method.

\subsection{Multi-GNSS Pseudorange Constraint}

To obtain the absolute position, we use a multi-GNSS pseudorange constraint. In contrast to other studies, we simultaneously use multiple GNSS observations such as GLONASS, Galileo, and BeiDou. The GNSS pseudorange measurement represents the distance between each GNSS satellite and the antenna. However, the GNSS pseudorange includes the receiver clock errors as well as the errors related to signal propagation. In this study, the pseudorange error is computed as:

\begin{equation}
e_{\mathrm{pr}, i}^{k}={H}_{i}^{k} {X}_{i}-\left(\rho_{i}^{k}-r_{i}^{k}+\delta T_{i}^{k}-I_{i}^{k}-T_{i}^{k}\right)
\end{equation}

In the aforementioned equation, $\rho_{i}^{k}$ is the observed GNSS psudorange. $r_{i}^{k}$ is the geometric satellite-to-receiver distance, which is calculated using initial node. $\delta T_{i}^{k}$ as the clock bias of the satellite, $I_{i}^{k}$ is the ionospheric delay and $T_{i}^{k}$ is the tropospheric delay. The clock bias $\delta T_{i}^{k}$ of the $k$-th satellite can be calculated from the broadcasted clock error information. The tropospheric delay $T_{i}^{k}$ is compensated using the Saastamoinen and Hopfield models \cite{gnss}. The ionospheric delay $I_{i}^{k}$ is compensated using the Klobuchar model \cite{gnss}. Here, measurement matrix ${H}_{i}^{k}$ can be formulated as:

\begin{equation}
{H}_{i}^{k}=\left[\begin{array}{llllllll}
{u_{x,i}^{k}} & {u_{y,{i}}^{k}} & {u_{z,{i}}^{k}} & {1} & {\delta_{\mathrm{GLO}, i}^{k}} & {\delta_{\mathrm{GAL, i}}^{k}} & {\delta_{\mathrm{BDS}, i}^{k}}
\end{array}\right]
\end{equation}

where $u_{x,i}^{k}$, $u_{y,i}^{k}$, $u_{z,i}^{k}$ is the unit line-of-sight vector from the receiver to the satellite $k$; and $\delta_{\mathrm{GLO}, i}^{k}$, $\delta_{\mathrm{GAL, i}}^{k}$, and $\delta_{\mathrm{BDS}, i}^{k}$ are equal to 1 when the $k$-th GNSS measurement is GLONASS, Galileo, or BeiDou, respectively. The minimized error of the GNSS pseudorange constraint is calculated as:

\begin{equation}
\left\|e_{\mathrm{pr}, i}^{k}\right\|_{\Omega_{\mathrm{pr}, i}^{k}}=e_{\mathrm{pr}, i}^{k} \hspace{1pt} \Omega_{\mathrm{pr}, i}^{k} \hspace{1pt} e_{\mathrm{pr}, i}^{k}
\end{equation}

where $\Omega_{\mathrm{pr},i}^{k}$ represents the information matrix of the GNSS pseudorange computed from the variance of pseudorange measurement. The GNSS pseudorange variance is determined as a function of the satellite elevation angle. The absolute 3D position in the ECEF coordinate can be determined to use the pseudorange constraint.

\subsection{Optimization}

We add all the constraints into the pose graph. The least squares solution of the factor graph with nodes and edges as described previously can be formulated as:

\begin{equation}
\begin{aligned} 
\widehat{\boldsymbol{X}}=\underset{X}{\operatorname{argmin}} \sum_{i}\left\|e_{\mathrm{vel}, i}\right\|_{\Omega_{\mathrm{vel}, i}}^{2}+\sum_{j}\left\|e_{\mathrm{tr}, j}\right\|_{\Omega_{ \mathrm{tr}, j}}^{2} \\ +\sum_{i} \sum_{k}\left\|e_{\mathrm{pr}, i}^{k}\right\|_{\Omega_{\mathrm{pr}, i}^{k}}^{2}
\end{aligned}
\end{equation}

The objective of localization is to find the optimal state variables that guarantee the minimized aforementioned error function. We use the Dogleg optimizer \cite{tuto} to obtain the optimal solution of the aforementioned error function. Finally, we can estimate the accurate vehicle's trajectory. The proposed method using TR-RTK-GNSS facilitates the realization of centimeter precision in applications without using any additional sensors and data from a GNSS base station.

\section{Experiments}

\subsection{Setup}
To confirm the effectiveness of the proposed technique, we conducted three kinematic positioning tests using an unmanned aerial vehicle (UAV). Currently, UAVs have been widely investigated for a variety of applications including precise 3D mapping. In direct 3D mapping, the UAV's precise position and attitude are necessary for constructing 3D maps. In general, 3D position estimation is based on a dual-frequency RTK-GNSS and a high-grade inertial measurement unit. If the precise UAV 3D trajectory can be estimated by using only a stand-alone GNSS receiver, the total cost of 3D mapping using the UAV can be reduced, and the 3D mapping will be more efficient. 
Some datasets exist for the benchmark of robot localization, such as the KITTI dataset \cite{kitti} and smartLoc project \cite{smaloc}. However, these datasets do not contain full GNSS raw data (e.g., pseudorange, carrier phase, and Doppler). We conducted three kinematic positioning tests to evaluate the proposed method because the proposed method uses GNSS raw data extensively. Fig. 3 shows the UAV equipped with a GNSS receiver and antenna. We used a UAV (DJI MATRICE 600 PRO) equipped with a sensor to collect data. A low-cost single-frequency GNSS antenna and u-blox NEO M8T receiver were used to obtain the GNSS data. The GNSS data were obtained at a rate of 1 Hz. We used the Applanix APX-15 UAV \cite{apx}, which is a position and attitude estimation system based on the integration of RTK-GNSS and high-grade IMU, to obtain the ground truth. APX-15 is typically used to obtain a reference position for small UAVs. The accuracy of the absolute position estimation is 2 cm according to the catalog specification. Therefore, it is accurate enough to be used to obtain the ground truth to compare the proposed methods. The conventional baseline to compare the performance is the general GNSS/IMU navigation solution \cite{gpsins}, which is typically used in the UAV navigation. We also compare the performance using conventional GNSS single point positioning results.

We used the GTSAM \cite{gtsam} for graph optimization backend, and RTKLIB \cite{rtklib} for GNSS general computation. The proposed method is implemented using MATLAB, and the performance evaluation was conducted in post-processing. 

The three tests were conducted in different locations in Japan. Fig. 4 shows the flight routes of the UAV. The UAV autonomously flew for approximately 200 m at speeds of less than 2.5 m/s. In these experiments, no obstacles were present around the GNSS antenna; however, the carrier phase cycle slips occurred occasionally in the low elevation angle satellite.

\subsection{Graph Construction}
Fig. 5 shows the pose graph constructed using the proposed method. The initial nodes of the graph (plotted as blue points in Fig. 5) were computed from the GNSS Doppler velocity. The red lines in Fig. 5 indicate the loop closing edge added to the graph by the proposed TR-RTK-GNSS. The distant nodes are connected by the constraints computed using the TR-RTK-GNSS. The figure shows that, TR-RTK-GNSS cannot generate constraints between long time separated nodes because the satellite clock bias error increases with time in TR-RTK-GNSS. Fig. 6 shows the histogram of the time differences of the fixed solution computed from the TR-RTK-GNSS technique. The maximum time difference was 95 s in these tests, which was the limitation for the construction of the loop-closing constraint by the proposed TR-RTK-GNSS technique. In other words, we can reduce the accumulated errors of the Doppler velocity within about 100 seconds by graph optimization.

\begin{figure}[t]
   \centering
   \includegraphics[width=85mm]{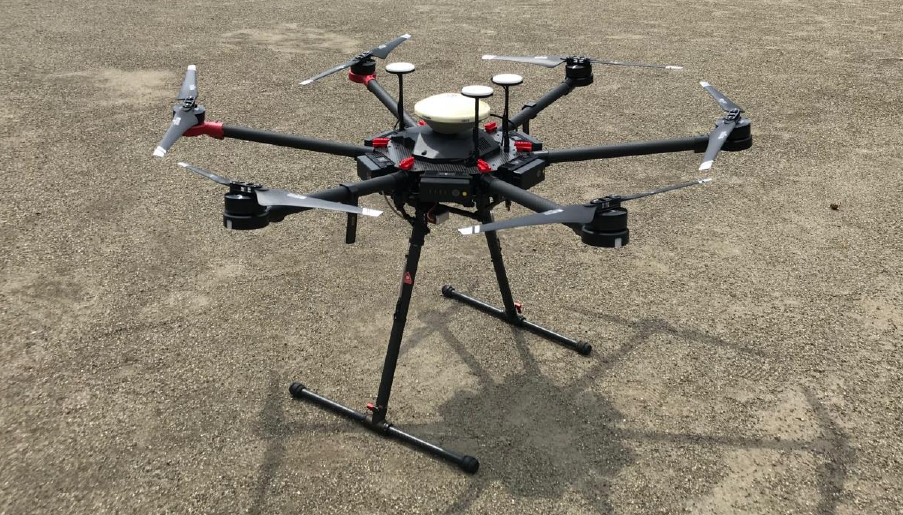} 
   \caption{UAV equipped with GNSS receiver and antenna. }
   \label{fig3}
\end{figure}

\begin{figure*}[t!]
   \centering
   \includegraphics[width=180mm]{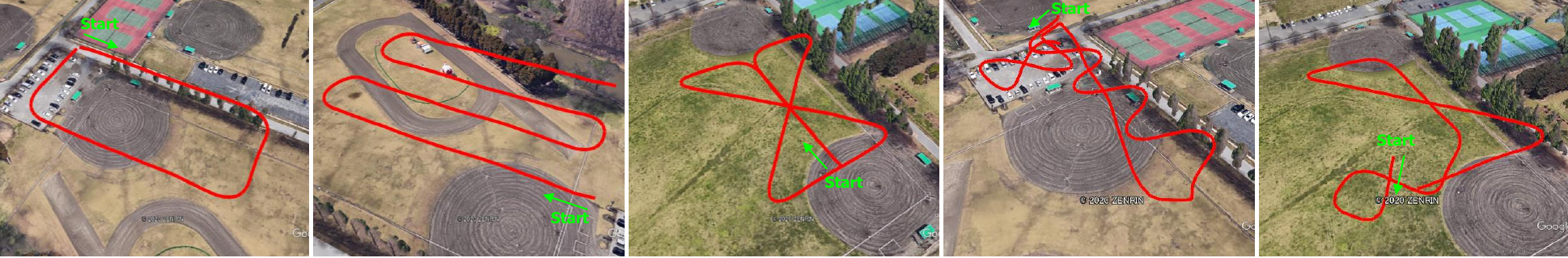} 
   \caption{Test environment and UAV trajectory (red line). These tests were conducted in open-sky environments. The flight altitude is approximately 35 m, and flight speed is 2.5 m/s.}
   \label{fig4}
\end{figure*}

\begin{figure*}[t!]
   \centering
   \includegraphics[width=180mm]{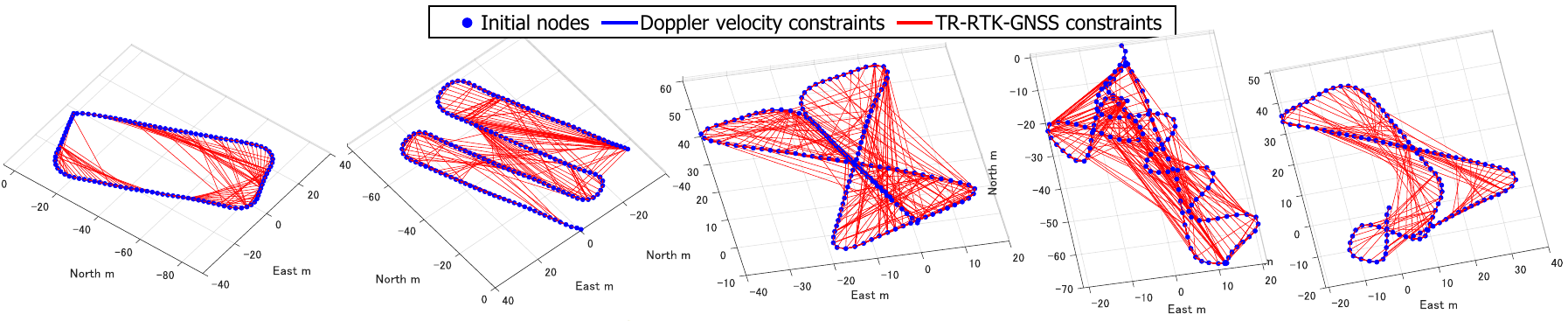} 
   \caption{Pose graph of the proposed method for the UAV experiment. The blue points indicate the initial nodes generated from GNSS Doppler velocity. The red lines indicate the loop-closing edge computed from TR-RTK-GNSS. The constraints of TR-RTK-GNSS connect the distant nodes.}
   \label{fig5}
\end{figure*}

\begin{figure*}[t!]
   \centering
   \includegraphics[width=180mm]{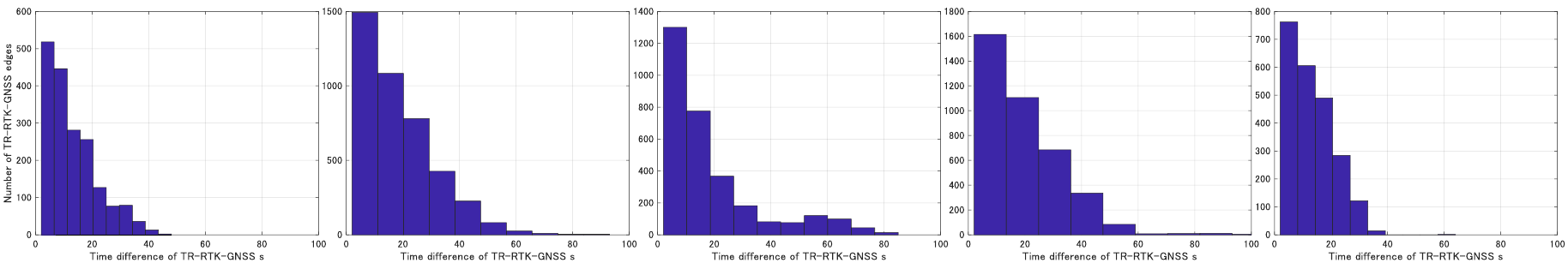} 
   \caption{Histogram of the time differences of the fixed solution generated by TR-RTK-GNSS. TR-RTK-GNSS can be adopted for generating constraints within approximately 100 s.}
   \label{fig6}
\end{figure*}

\begin{figure*}[t!]
   \centering
   \includegraphics[width=180mm]{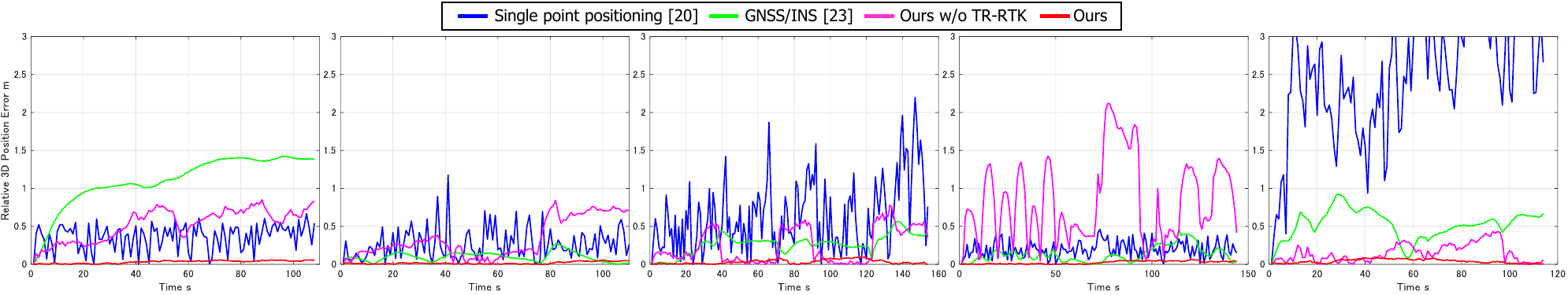} 
   \caption{3D relative localization errors of each method. The blue line indicates the conventional single point positioning computed by \cite{rtklib}. The green line is GNSS/INS solutions by \cite{gpsins}. The pink line is the proposed method without using TR-RTK-GNSS. If we use TR-RTK-GNSS (red line), the accumulated error can be reduced.}
   \label{fig7}
\end{figure*}

\begin{figure*}[t!]
   \centering
   \includegraphics[width=160mm]{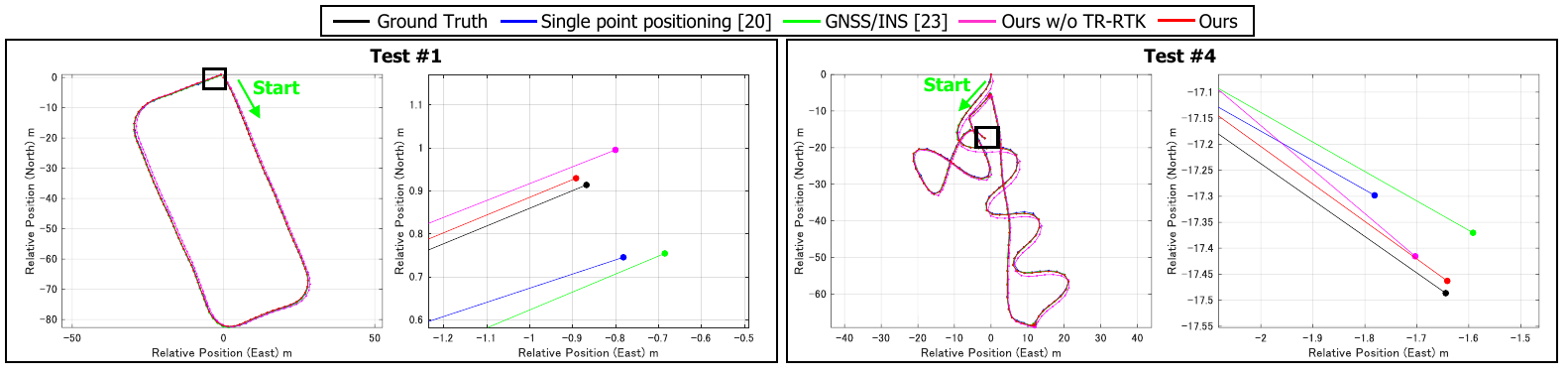} 
   \caption{Examples of estimated trajectory of each method (left: Test \#1 and right: Test \#4) and the close-up image of the last point. The black line indicates the ground truth computed by \cite{apx}. The proposed method (red line) is the closest to the ground truth.}
   \label{fig8}
\end{figure*}

\begin{table*}[thb]
   \caption{Comparison of the position estimation accuracy of the proposed method and the existing methods}
   \label{tab1}
   \resizebox{\textwidth}{!}{%
   \begin{tabular}{c|ccc|ccc|ccc|ccc|ccc}
   \hline
   \multirow{2}{*}{Method}          & \multicolumn{3}{c|}{Test   \#1}                                                & \multicolumn{3}{c|}{Test   \#2}                                                & \multicolumn{3}{c|}{Test   \#3}                                                & \multicolumn{3}{c|}{Test   \#4}                                                & \multicolumn{3}{c}{Test   \#5}                                                 \\ \cline{2-16} 
                                    & RPE m          & \begin{tabular}[c]{@{}c@{}}Max. \\ RPE m\end{tabular} & APE m & RPE m          & \begin{tabular}[c]{@{}c@{}}Max. \\ RPE m\end{tabular} & APE m & RPE m          & \begin{tabular}[c]{@{}c@{}}Max. \\ RPE m\end{tabular} & APE m & RPE m          & \begin{tabular}[c]{@{}c@{}}Max. \\ RPE m\end{tabular} & APE m & RPE m          & \begin{tabular}[c]{@{}c@{}}Max. \\ RPE m\end{tabular} & APE m \\ \hline
   Single point positioning{[}20{]} & 0.399          & 0.668                                                 & 2.411 & 0.341          & 1.176                                                 & 2.674 & 0.791          & 2.198                                                 & 2.539 & 2.655          & 4.328                                                 & 2.2   & 0.223          & 0.461                                                 & 0.342 \\
   GNSS/INS{[}23{]}                 & 1.153          & 1.424                                                 & 1.482 & 0.109          & 0.268                                                 & 0.847 & 0.295          & 0.564                                                 & 0.887 & 0.558          & 0.922                                                 & 0.978 & 0.165          & 0.462                                                 & 0.824 \\
   Ours w/o TR-RTK                  & 0.568          & 0.851                                                 & 1.899 & 0.563          & 1.338                                                 & 0.965 & 0.334          & 0.778                                                 & 1.186 & 0.191          & 0.433                                                 & 0.651 & 0.995          & 2.12                                                  & 1.146 \\
   Ours                             & \textbf{0.037} & \textbf{0.059}                                        & 1.823 & \textbf{0.026} & \textbf{0.056}                                        & 0.592 & \textbf{0.044} & \textbf{0.096}                                        & 0.996 & \textbf{0.049} & \textbf{0.086}                                        & 0.698 & \textbf{0.031} & \textbf{0.062}                                        & 0.605
   \end{tabular}%
   }
\end{table*}

\subsection{Results}
In order to evaluate the accuracy of the proposed method, we use two metrics, relative position error (RPE) and absolute position error (APE). The RPE represents the relative error of the flight trajectory from the start point. Estimating the accurate trajectory is very important for the 3D measurement by UAVs. Fig. 7 shows the 3D RPE of each method. Here, "Ours w/o TR-RTK" in Fig. 7 means that we use only Doppler velocity constraints and multi-GNSS pseudorange constraints for graph optimization. We can see that the single point positioning has errors in several meters. In the conventional GNSS/INS and "Ours w/o TR-RTK", the error varies with the time. However, the proposed method can reduce the positioning errors, and the RPE is close to zero. This result shows that the TR-RTK-GNSS constraint allows us to estimate the UAV's 3D flight trajectory with very high accuracy. Fig. 8 shows the estimated trajectory (2D representation) of each method in Test \#1 and Test \#4. The proposed method is closest to the ground truth. Table 1 shows the 3D RPE, the maximum RPE, and the 3D APE of each method. The 3D RPE of the estimated trajectory of the proposed method was 3.7, 2.6, 4.4, 4.9, and 3.1 cm for the five flight tests, respectively. The maximum RPE of the proposed method in the five flight tests was 9.6 cm. However, for APE, the proposed method has almost the same or slightly lower errors than other methods. This means that, the 3D position estimated by the proposed method has a constant bias error. However, the proposed method estimated an exact centimeter-level trajectory using only a stand-alone GNSS receiver without using a GNSS base station, IMU, or other sensors.

\subsection{Discussion}
The proposed method can estimate the vehicle's relative position with an accuracy of few centimeters and with meter-level 3D bias error in geospatial coordinates. This bias error can be compensated by one additional constraint, namely prior knowledge of the initial position. From the tests, the TR-RTK-GNSS constraint can be generated when the time difference is less than approximately 100 s. In other words, the loop of the pose graph can be detected and closed within approximately 100 s. For this reason, the UAV flight path affects the final accuracy of the proposed method. The proposed method is more effective with a flight that has more turns than a straight flight, and it can close the loop in a short time.

\section{Conclusion and Future Work}

\subsection{Conclusion}
In this study, pose-graph optimization using GNSS was proposed to estimate the precise trajectory of vehicles. A loop closure of the pose graph was generated from TR-RTK-GNSS technique, which is a method of implementing a precise carrier phase-based time-differential GNSS with a single low-cost GNSS receiver. Unlike conventional RTK-GNSS, we can directly compute the vehicle’s relative position using only a stand-alone GNSS receiver. The initial pose graph was generated from the accumulated velocity that was computed based on GNSS Doppler measurements. To eliminate the accumulated error of velocity, we used TR-RTK-GNSS for the loop closure in graph-based optimization framework. Thus, our study showed that, we could estimate the vehicle trajectory with centimeter accuracy using only a stand-alone GNSS receiver. Test results revealed that, the proposed technique is effective at estimating the precise vehicle location.

\subsection{Future Work}
In future research, we will develop the existing method to take into account the time difference limitation of the TR-RTK-GNSS technique. We will also apply the proposed method to localize vehicles in urban environments where GNSS multipath occurs. To detect and exclude multipath errors, we will use a robust optimization technique. In addition, we will extend the proposed technique to corroborate localization of multiple robots.

\ifCLASSOPTIONcaptionsoff
  \newpage
\fi



\bibliographystyle{IEEEtran}
\bibliography{IEEEabrv,ra-l}

\end{document}